\documentclass[journal]{IEEEtran}
\usepackage{lineno}
\usepackage{bbm}
\usepackage{bm}
\usepackage{amsfonts}
\usepackage{amsmath}
\usepackage{mathrsfs}
\usepackage{amssymb}
\usepackage{enumerate}
\usepackage{graphicx}
\usepackage{subfigure}
\usepackage{booktabs}
\usepackage{graphicx}
\usepackage{setspace}
\usepackage{algorithm}
\usepackage{algorithmic}
\usepackage{setspace}
\usepackage{multirow}
\usepackage{array}
\usepackage{color}
\usepackage{cite}
\usepackage{geometry}
\usepackage{hyperref}
\usepackage{bbding}
\usepackage{pifont}
\usepackage{ulem}
\ifCLASSINFOpdf
\else
\fi
\begin{document}
\title{CLIP-Driven Semantic Discovery Network for Visible-Infrared Person Re-Identification}
\author{Xiaoyan~Yu, Neng~Dong, Liehuang Zhu, Hao Peng, Dapeng Tao
\thanks{This research was supported by the National Natural Science Foundation of China (Nos. 62276120, 61966021).}
\thanks{X. Yu is with the School of Computer Science and Technology, Beijing Institute of Technology, Beijing 100081, China  (e-mail: xiaoyan.yu@bit.edu.cn).}
\thanks{N. Dong is with the School of Computer Science and Engineering, Nanjing University of Science and Technology, Nanjing 210094, China  (e-mail: neng.dong@njust.edu.cn).}
\thanks{L. Zhu is with with the School of Computer Science and Technology, Beijing Institute of Technology, Beijing 100081, China (e-mail: liehuangz@bit.edu.cn).}
\thanks{H. Peng is with the School of Cyber Science and Technology, Beihang University, Beijing 100191, China (e-mail: penghao@buaa.edu.cn).}
\thanks{Dapeng Tao is with the Fist Laboratory, School of Information Science and Engineering, Yunnan University, Kunming 650091, China (e-mail: dapeng.tao@gmail.com).}
}
\markboth{SUBMISSION OF IEEE Transactions on Multimedia, 2024}%
{Shell \MakeLowercase{\textit{et al.}}: Bare Demo of IEEEtran.cls for IEEE Journals}
\maketitle
\begin{abstract}
Visible-infrared person re-identification (VIReID) primarily deals with matching identities across person images from different modalities. Due to the modality gap between visible and infrared images, cross-modality identity matching poses significant challenges.  Recognizing that high-level semantics of pedestrian appearance, such as gender, shape, and clothing style, remain consistent across modalities, this paper intends to bridge the modality gap by infusing visual features with high-level semantics. Given the capability of CLIP to sense high-level semantic information corresponding to visual representations, we explore the application of CLIP within the domain of VIReID. Consequently, we propose a CLIP-Driven Semantic Discovery Network (CSDN) that consists of Modality-specific Prompt Learner, Semantic Information Integration (SII), and High-level Semantic Embedding (HSE). Specifically, considering the diversity stemming from modality discrepancies in language descriptions, we devise bimodal learnable text tokens to capture modality-private semantic information for visible and infrared images, respectively. Additionally, acknowledging the complementary nature of semantic details across different modalities, we integrate text features from the bimodal language descriptions to achieve comprehensive semantics. Finally, we establish a connection between the integrated text features and the visual features across modalities. This process embed rich high-level semantic information into visual representations, thereby promoting the modality invariance of visual representations. The effectiveness and superiority of our proposed CSDN over existing methods have been substantiated through experimental evaluations on multiple widely used benchmarks. The code will be released at \url{https://github.com/nengdong96/CSDN}.

\end{abstract}
\IEEEpeerreviewmaketitle

\section{Introduction}\IEEEPARstart{P}{erson} Re-Identification (ReID) aims to retrieve pedestrian images belonging to the same identity from non-overlapping cameras. This task holds significant importance for public security maintenance, including criminal apprehension, locating missing individuals, and more. The study of person ReID has extended over numerous years and recently yielded commendable achievements \cite{BAGTRICKS, AGW, HDSA, ETNDNet}. However, the majority of mainstream algorithms struggle to accommodate the demands of 24-hour intelligent surveillance since they mainly focus on retrieving visible images procured from RGB cameras. In real-world scenarios, surveillance cameras automatically capture infrared images during nighttime, which have a distinct wavelength range from the visible ones. Employing the single visible modality ReID framework for retrieving infrared images would lead to a substantial decline in performance. To this end, the cross-modality Visible-Infrared person Re-Identification (VIReID) \cite{VIREID,IGBS,LMIT} has been proposed, which retrieves images with the same identity as the provided visible (infrared) query from a gallery containing infrared (visible) images.

\begin{figure}[t!]
	\centering
	\includegraphics{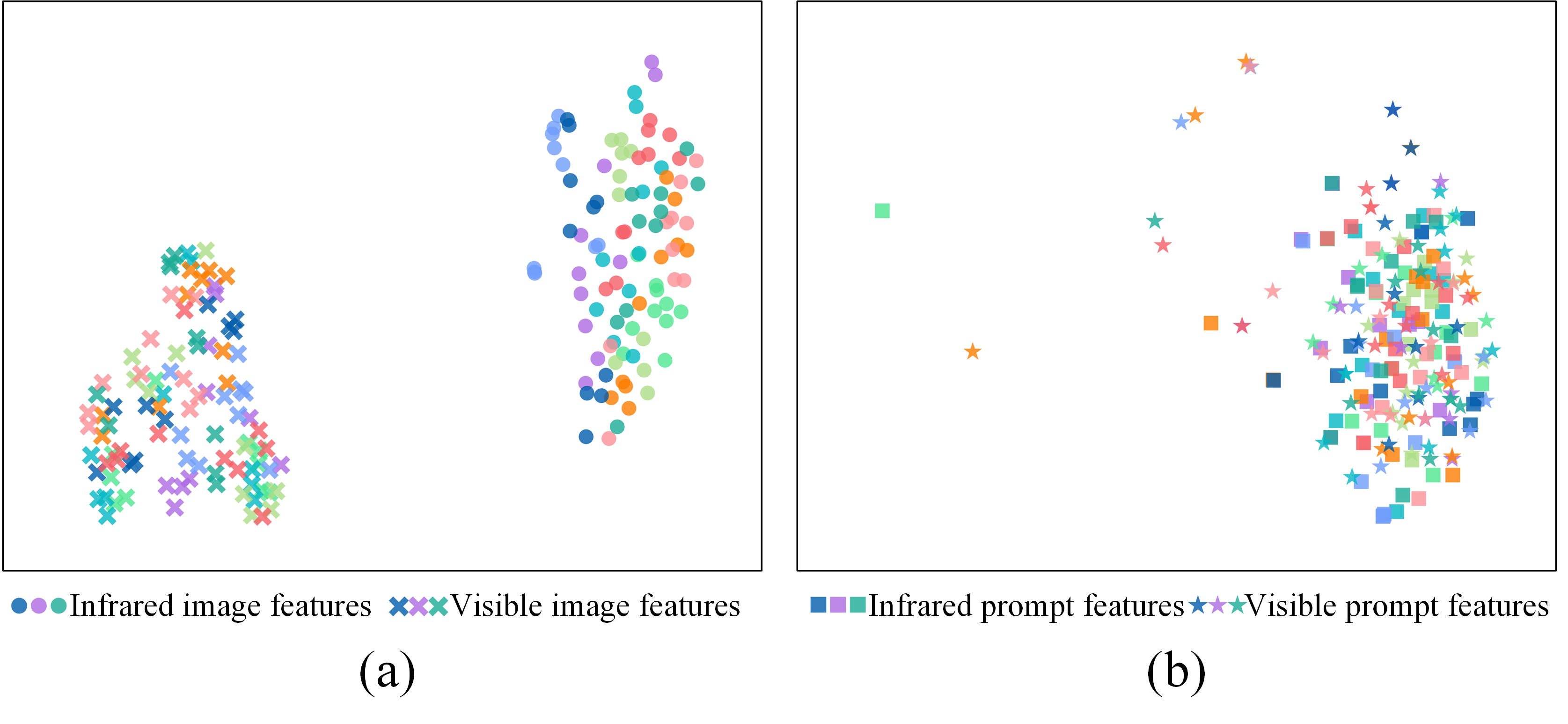}
	\caption{The core motivation of this paper. The image features of visible and infrared modalities exhibit significant modality discrepancies (see (a)), while their corresponding text features reveal no such disparities (see (b)). Consequently, employing textual features as a bridge to align visual representations across diverse modalities is deemed feasible.}
	\label{Fig:1}
\end{figure}

As a cross-modality image retrieval task, the critical challenge of VIReID lies in mitigating the substantial modality gap arising from heterogeneous data sources. Existing methods can be broadly categorized into two groups: generative-based algorithms \cite{D2RL, AlignGAN, HiCMD, JSIA} and non-generative-based algorithms \cite{cmGAN, MSR, SDL, DFE}. The former employs Generative Adversarial Networks (GAN) \cite{GAN} to transfer images from one modality to another or to create intermediate modality images for training, bridging the discrepancy between two modalities at the image level. However, the presence of noise interference hinders the training stability, making it challenging to ensure the quality of generated images. Without requiring the generation process, the latter is dedicated to formulating network architectures and metric functions for modality alignment at the feature level. Despite their effectiveness, existing frameworks train models solely with images, while visual contents learned exclusively under the supervision of images lack high-level semantic information \cite{CLIP-ReID}, making modality alignment still challenging.

Recently, the paradigm of visual-language learning has garnered significant attention owing to its capacity for learning semantically rich visual representations. Contrastive Language-Image Pre-training (CLIP) \cite{CLIP}, a prominent cross-modal pre-training model, has attained remarkable success and been applied to various downstream visual tasks \cite{CRIS, DenseCLIP, CFine}. In the community of person ReID, recent studies \cite{CLIP-ReID, RGAN, VL} have demonstrated that employing CLIP to bridge visual content with its corresponding language description enables the model to sense high-level semantics related to target pedestrians. However, these investigations predominantly concentrate on single-modality ReID, with the application of CLIP in VIReID remaining unexplored. In particular, according to our observations illustrated in Figiue \ref{Fig:1}(a), high-level semantic information corresponding to images across different modalities, such as gender, hairstyle, and shape, do not exhibit significant modality discrepancies. Therefore, the core motivation of this paper lies in adeptly adapting CLIP to the VIReID task, leveraging semantic information as a bridge for aligning the visual representations of visible and infrared images.

One of the simplest strategies to harness the potential of CLIP is by substituting the backbone with CLIP's pre-trained model. However, as elucidated in CLIP-ReID \cite{CLIP-ReID}, this method falls short of fully exploiting the potent capabilities inherent in CLIP. Consequently, inspired by CLIP-ReID, it is feasible to introduce a learning paradigm of the prompt learner to generate modality-shared natural language descriptions for pairs of cross-modality pedestrians. Subsequently, these generated natural language descriptions are employed to extract text features with high-level semantic information, guiding the learning and alignment of visual features. However, we contend that modality-shared language descriptions may be somewhat inadequate for VIReID learning. One primary reason is that descriptions of images across different modalities may emphasize different aspects. For instance, descriptions of visible images may focus on pedestrian clothing, while descriptions of infrared images may center on pedestrian shapes. Consequently, the modality-shared description may contain less high-level semantic information due to the loss of modality-specific details. Additionally, language descriptions corresponding to different modality images may be complementary. If we can organically integrate the two, more nuanced and comprehensive semantic information can be acquired, thereby facilitating a more effective alignment of the visual representations.

The preceding discussion motivates the development of an effective methodology in this paper for applying CLIP to VIReID tasks, termed CLIP-driven Semantic Discovery Network (CSDN). To be specific, as illustrated in Figure \ref{Fig:2}, we first design a learning paradigm named Modality-specific Prompt Learner (MsPL), whose objective is to generate bimodal natural language descriptions for each identity. These two language descriptions contain the semantic information of pedestrians in visible modality and infrared modality respectively. Considering the complementary nature of semantic details across the two modalities for the same pedestrian, we further develop a Semantic Information Integration (SII) module, in which an attention fusion mechanism is incorporated to merge the text features derived from bimodal natural language descriptions. The resultant integrated text features encompass the semantic details of pedestrians in both visible and infrared modalities, providing a more comprehensive characterization of individuals. Finally, to learn semantically rich visual representations that facilitate modality alignment, we devise a High-level Semantic Embedding (HSE) module to establish the connection between the rich textual features and visual features of pedestrian images.

Our main contributions are summarized as follows:
\begin{itemize}
\item \textbf{Fresh Perspective}. To address the hindrance brought by modality gap in cross-modality matching, this paper proposes to inject the semantic information carried by high-level language descriptions into visual features to improve the modality invariance of visual features. To the best of our knowledge, this is the first time that the semantic information carried by language descriptions is used to improve the modality robustness of visual features. This is a new idea independent of existing VIReID methods.

\item \textbf{Distinctive Proposition}. We explore the application of CLIP in VIReID. We design MsPL to employ two independent paths for generating high-level language descriptions for infrared and visible images, effectively preserving the complementary semantic information between bimodal images. Additionally, our proposed SII integrates the generated high-level language descriptions, guiding the injection of semantic information, the lack of modality-specific information is compensated, and the modality discrepancy is effectively reduced.

\item \textbf{Excellent performance}. We validate the performance of our method on two widely used datasets. The experimental results consistently surpassed those of currently favored methods, which demonstrates the effectiveness and superiority against state-of-the-art methods.
\end{itemize}

The rest of the content is structured as follows. Section \uppercase\expandafter{\romannumeral2} reviews some related work. Section \uppercase\expandafter{\romannumeral3} introduces the proposed method in detail. Section \uppercase\expandafter{\romannumeral3} presents extensive experiments to verify the effectiveness of the proposed method. Section \uppercase\expandafter{\romannumeral4} summarizes the proposed method and draws some conclusions.

\section{Related Work}
\subsection{Single-Modality Person ReID}
Typical person ReID refers to single-modality person ReID, wherein the objective is to retrieve visible images of interested pedestrians. The principal challenge inherent in single-modality ReID is to alleviate the adverse impact on recognition performance resulting from variations in camera views, background interference, posture changes, and occlusion noise. Yi \emph{et al}. \cite{DML} developed a Siamese network that pulls the same pedestrian from different views close and pushes different pedestrians from the same view away, facilitating the learning of view-invariant representations. Dong \emph{et al}. \cite{MVIIP} devised a multi-view information integration and propagation mechanism to excavate comprehensive information robust to camera views and occlusion. To protect the ReID system from background perturbation, Song \emph{et al}. \cite{MGCA} employed pedestrians' masks and proposed an attention module to suppress background noise. Tian \emph{et al}. \cite{EBB} randomly replaced the background of pedestrian images and thus eliminated background bias with the guidance of person-region. Moreover, Qian \emph{et al}. \cite{PNGAN} introduced a Generative Adversarial Network \cite{GAN} to formulate a pose-normalized image generation model, enabling the ReID system to learn the deep feature free of the influence of pose variations. Ge \emph{et al}. \cite{FDGAN} designed multiple discriminators and a novel same-pose loss to distinguish identity-related and pose-unrelated representations. Currently, diverse algorithms for robust pedestrian information mining have emerged \cite{AANet, DIF, TALMVR}, yielding satisfactory recognition performance in single-modality person ReID. However, meeting the demands of real scenarios proves challenging as the given target pedestrian image (query) or the sample set to be retrieved (gallery) may be an infrared image captured at night. In this paper, the developed CSDN is specifically tailored for cross-modality pedestrian image retrieval, offering contributions for practical applications of person ReID.

\subsection{Cross-Modality Person Re-ID}
Cross-modality person ReID refers to visible-infrared person ReID (VIReID) \cite{VIREID}, aiming to retrieve infrared images with consistent identity given a visible pedestrian image, and vice versa. Existing VIReID methods can be broadly categorized into generative-based and non-generative-based algorithms.

Generative-based methods transfer visible (infrared) images to infrared (visible) modality style or generate intermediate modality images that contain both visible and infrared styles to train the model, alleviating modality differences at the image level. For example, Wang \emph{et al}. \cite{AlignGAN} devised a pixel alignment module converting real visible images into synthetic infrared ones, reducing cross-modality variations. Choi \emph{et al}. \cite{HiCMD} employed a decoupling-generation strategy to disentangle identity-discriminative and identity-excluded factors. Considering the inherently challenging infrared-to-visible generation process due to less information in infrared images, Li \emph{et al}. \cite{X-modality} introduced auxiliary $X$ modality images to reconcile both the infrared and visible modalities, making cross-modality learning easier. However, these approaches necessitate a meticulously designed generation strategy to prevent mode collapse and the quality of the generated images proves challenging due to the presence of noise interference. Despite recent researches \cite{TS-GAN, CECNet, RBDF} dedicated to addressing this issue and achieving commendable results, the quality of the generated images remains unsatisfactory. In contrast, the proposed CSDN circumvents these issues by refraining from the generation process.

Non-generative-based methods concentrate on designing suitable network structures and effective metric functions for achieving modality alignment at the feature level. Specifically, Wu \emph{et al}. \cite{FMSP} published the first large-scale cross-modality pedestrian dataset and proposed a one-stream network with deep zero-padding. Hao \emph{et al}. \cite{HSME} designed a two-stream network where domain-specific layers extract modality-private features, and the domain-shared layer is tasked with mining modality-public information. To weaken the destruction of bad examples on pairwise distances, Liu \emph{et al}. \cite{PSEHCT} devised a hetero-center triplet loss, ensuring the compactness of intra-class features and the distinguishable property of inter-class features. Furthermore, Ling \emph{et al}. \cite{MCSL} formulated a multi-constraint similarity learning method to comprehensively explore the relationship among cross-modality informative pairs. Recently, some state-of-the-art methods \cite{DDAG, SCET, NFS, CIMA} have further boosted the performance of VIReID. However, these studies ignore the high-level semantic information unaffected by modality, restricting the model's capability to align features of visible and infrared images. In this paper, our CSDN endeavors to learn a rich language description to supervise the learning of semantically visual representations.

\subsection{Vision-Language Learning}
The vision-language learning paradigms \cite{NTS, FILIP, UNITER} have gained considerable popularity in recent years. Contrastive Language-Image Pre-training (CLIP) \cite{CLIP}, a representative vision-language learning model, establishes a connection between natural language and visual content through the similarity constraint of image-text pair. Enjoying CLIP's powerful capability for mining semantic information of visual representations, diverse strategies (e.g., CoOp \cite{COOP}, Adapter \cite{ADAPTER}) have been introduced to extend the applicability of CLIP to a series of downstream computer vision tasks, including image classification \cite{CHILS}, object detection \cite{CLIPOD}, semantic segmentation \cite{ZEGCLIP}, and etc.

Recently, the person ReID community has directed its attention towards CLIP \cite{CLIP-ReID, RGAN, VL}, seeking to leverage its capabilities for advancing the field further. Li \emph{et al}. pioneered the CLIP-ReID model \cite{CLIP-ReID}, acquiring language descriptions associated with pedestrian identity through the training of learnable prompts. Subsequently, the visual encoder is supervised by natural language to extract image features enriched with semantic information. However, the substantial potential of CLIP to promote VIReID learning has not been explored. In this paper, we initially define a CLIP-VIReID framework. Building on this, we further propose our CSDN, contributing to the mitigation of the modality discrepancy.

\begin{figure*}[t!]
	\centering
	\includegraphics[width=6.9in,height=4.4in]{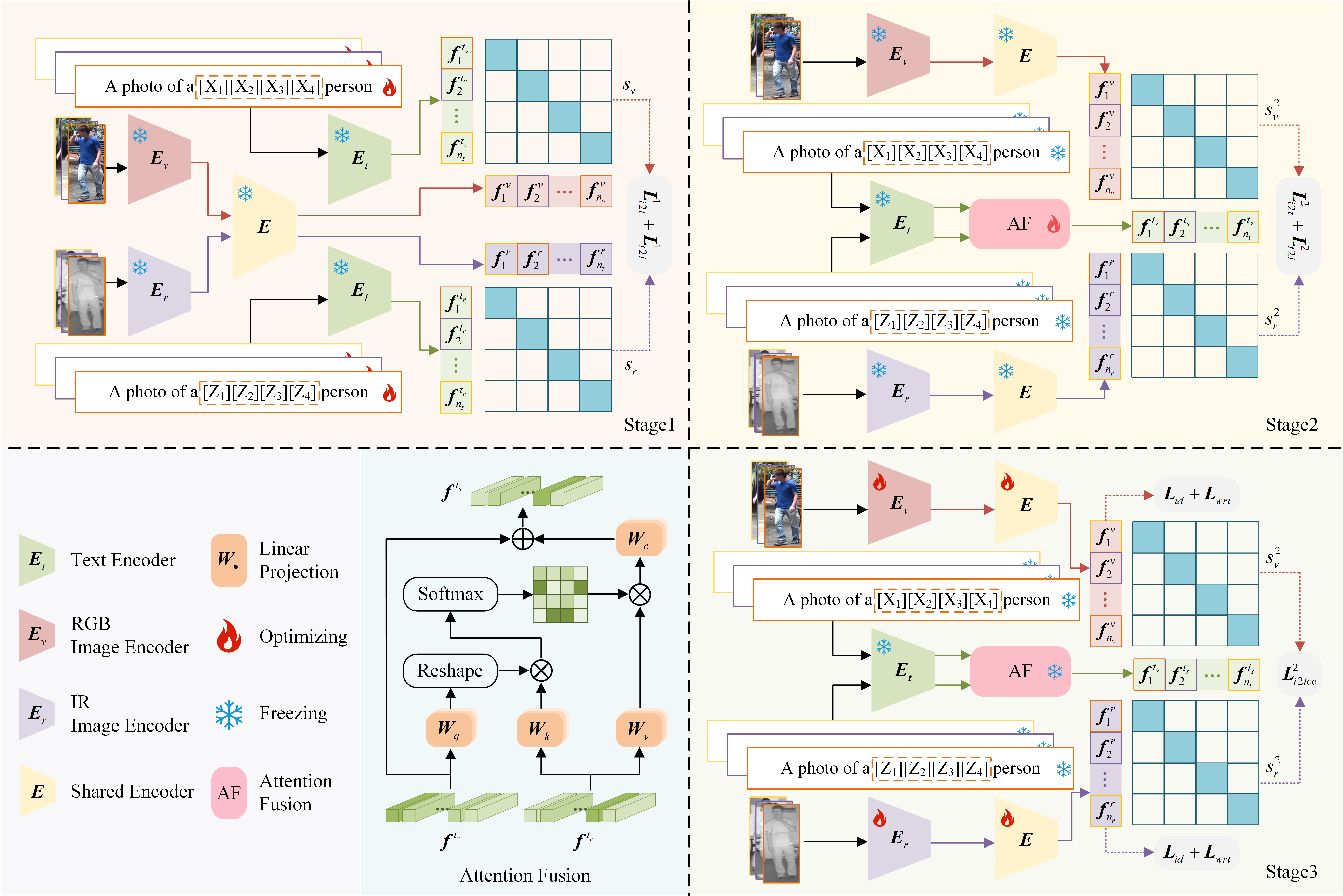}
	\caption{Overview of the proposed method. The entire learning process of our CSDN includes three stages. Stage 1 (MsPL): Given image samples across modalities, we design bimodal prompt learners to generate natural language descriptions corresponding to visible and infrared images respectively. Stage 2 (SII): Considering the semantic complementarity of descriptions in different modalities, we devise an attention fusion module to integrate their semantic details. Stage 3 (HSE): With the guidance of the integrated rich complementary semantics, we inject the semantic information into visual representations of visible and infrared images, promoting their modality invariance.}
	\label{Fig:2}
\end{figure*}

\section{The proposed methods} 

\subsection{Preliminaries}
Compared with single-modality ReID, VIReID is a challenging cross-modality identity-matching task due to the large modality differences between visible and infrared images. Let $\bm D=\{\bm D^{v}, \bm D^{r}, \bm Y \}$ denote the sample set, where $\bm D^{v}=\{\bm x_{i}^{v}\}_{i=1}^{N_{v}}$ represents $N_{v}$ visible images, $\bm D^{r}=\{\bm x_{i}^{r}\}_{i=1}^{N_{r}}$ describes $N_{r}$ infrared images, and $\bm Y=\{y_{i}\}_{i=1}^{N_{c}}$ is the label set. In existing VIReID studies, a standardized vision framework is employed to extract features for pedestrian identity matching. Specifically, the framework utilizes ResNet50 \cite{Resnet}, pre-trained on ImageNet \cite{Imagenet}, as the backbone. Two parallel shallow layers, denoted as $\bm E_{v}$ and $\bm E_{r}$, are utilized to capture modality-specific information. The remaining layers constitute the encoder $\bm E$, tasked with learning the modality-shared visual representation. The model is trained with the identity loss and weighted regularized triplet loss:

\begin{equation}
	\begin{aligned}
		L_{id}=-\frac{1}{n_{b}}\sum_{i=1}^{n_{b}} \bm q_{i}\log (\bm W (\bm f_{i}))
	\end{aligned},
 \label{Eq:1}
\end{equation}
where $\bm f_{i}=(\bm f_{i}^{v}, \bm f_{i}^{r})$ is the output feature of $\bm E$; $n_{b}$ represents the batch size; $\bm q_{i}$ is the one-hot vector of identity label $y_{i}$; $\bm W$ denotes the identity classifier.

\begin{equation}
	\begin{aligned}
		L_{wrt}=\frac{1}{n_{b}}\sum_{i=1}^{n_{b}}\log(1+\exp(\sum\nolimits_{ij}w^{p}_{ij}d^{p}_{ij}-\sum\nolimits_{ik}w^{n}_{ik}d^{n}_{ik}))
	\end{aligned},
 \label{Eq:2}
\end{equation}

\begin{equation}
	\begin{aligned}
		w_{ij}^{p}=\frac{\exp(d_{ij}^{p})}{\sum_{d_{ij}^{p}\in \mathcal{P}_{i}}\exp(d_{ij}^{p})}, w_{ik}^{n}=\frac{\exp(-d_{ik}^{n})}{\sum_{d_{ik}^{n}\in \mathcal{N}_{i}}\exp(-d_{ik}^{n})}
	\end{aligned}.
\end{equation}
where $j$ and $k$ are the indexes of the positive and negative samples corresponding to $\bm x_{i}$; $\mathcal{P}_{i}$ and $\mathcal{N}_{i}$ respectively represent the positive and negative sample sets corresponding to $x_{i}$ in a batch; $d_{ij}^{p}=\|\bm f_{i}- \bm f_{j}\|_{2}$ and $d_{ik}^{n}=\| \bm f_{i}- \bm f_{k}\|_{2}$ denote the Euclidean distance of the positive and negative sample pairs.

However, as discussed earlier, this conventional framework solely relies on images for model training and thus lacks the capacity to sense high-level semantic information conducive to mitigating modality differences. This limitation motivates us to inject the semantic details conveyed by high-level language descriptions into visual features, aiming to enhance the modality invariance of the visual representations.

\subsection{CLIP-VIReID}
\begin{figure*}[th!]
\centering
{\includegraphics[height=2.2in,width=6.8in,angle=0]{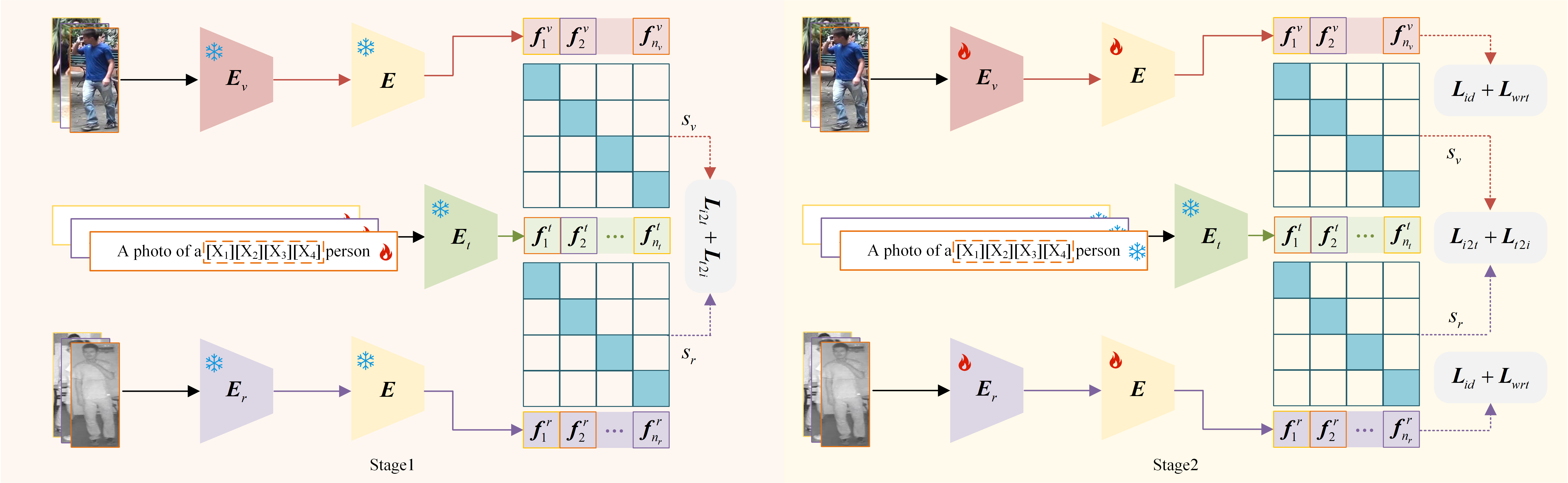}}
\caption{Our idea of applying CLIP on VIReID, and we name it CLIP-VIReID. Specifically, to harness CLIP’s potent capabilities, we build a learnable language description to acquire semantic information for pairs of cross-modality images. Subsequently, we employ the obtained semantics to establish connections of visual representations across different modalities.}
\label{Fig:3}
\end{figure*}

CLIP, a prominent multi-modal model trained on web-scale text-image pairs, excels at extracting semantically rich visual features by establishing a connection between natural language and vision. While employing its pre-trained visual encoder as the backbone of the VIReID framework is a logical option, it falls short of fully harnessing CLIP's potent capabilities to facilitate VIReID learning. Consequently, we explore a CLIP-VIReID model, as depicted in Figure \ref{Fig:3}.

Specifically, a crucial prerequisite for CLIP's capacity to sense high-level semantics is the availability of natural language descriptions corresponding to images. However, pedestrian images lack corresponding textual data. Drawing inspiration from CLIP-ReID \cite{CLIP-ReID}, we introduce a paradigm of the learnable prompt to generate language descriptions for pairs of cross-modality images. Define ``A photo of a [X]$_{1}$, [X]$_{2}$,\ldots, [X]$_{M}$, $[Cls]$ person'' as the learnable language description $\mathcal{T}_{i}$ of $(\bm x_{i}^{v}, \bm x_{i}^{r})$, where [X]$_{m}$ represents the trainable word token, $M$ is the total number of tokens, and $[Cls]$ denotes the identity label $y_{i}$ corresponding to $(\bm x_{i}^{v}, \bm x_{i}^{r})$. As shown in Figure \ref{Fig:3}, we utilize the CLIP pre-trained text encoder $\bm E_{t}$ to extract the features $\bm f_{i}^{t}$ from $\mathcal{T}_{i}$. To ensure one-to-one correspondence between $\mathcal{T}_{i}$ and $(\bm x_{i}^{v}, \bm x_{i}^{r})$, we employ image-to-text and text-to-image contrastive losses to optimize:

\begin{equation}
	\begin{aligned}
L_{i2t} =  &- \frac{1}{{{n_v}}}\sum\limits_{i = 1}^{{n_v}} {\log \frac{{\exp \left( {s\left( {\bm f_i^v, \bm f_{i}^{t}} \right)} \right)}}{{\sum\nolimits_{j = 1}^{{n_v}} {\exp \left( {s\left( {\bm f_i^v, \bm f_j^{{t}}} \right)} \right)} }}}  \\&- \frac{1}{{{n_r}}}\sum\limits_{i = 1}^{{n_r}} {\log \frac{{\exp \left( {s\left( {\bm f_i^r,\bm f_{i}^{{t}}} \right)} \right)}}{{\sum\nolimits_{j = 1}^{{n_r}} {\exp \left( {s\left( {\bm f_i^r, \bm f_j^{{t}}} \right)} \right)} }}} 
	\end{aligned},
 \label{Eq:4}
\end{equation}

\begin{equation}
	\begin{aligned}
	L_{t2i} =  - \frac{1}{{{n_v}}}\sum\limits_{i = 1}^{{n_v}} {\frac{1}{{\left| {P\left( {{y_i}} \right)} \right|}}\sum\limits_{{p_i} \in P\left( {{y_i}} \right)} {\log \frac{{\exp \left( {s\left( {\bm f_{{p_i}}^v,\bm f_{y_{i}}^{t}} \right)} \right)}}{{\sum\nolimits_{j = 1}^{{n_v}} {\exp \left( {s\left( {\bm f_j^v, \bm f_{y_{i}}^{t}} \right)} \right)} }}} }  \\ 
	{\rm{         }} - \frac{1}{{{n_r}}}\sum\limits_{i = 1}^{{n_r}} {\frac{1}{{\left| {P\left( {{y_i}} \right)} \right|}}\sum\limits_{{p_i} \in P\left( {{y_i}} \right)} {\log \frac{{\exp \left( {s\left( {\bm f_{{p_i}}^r, \bm f_{y_{i}}^{t}} \right)} \right)}}{{\sum\nolimits_{j = 1}^{{n_r}} {\exp \left( {s\left( {\bm f_j^r, \bm f_{y_{i}}^{t}} \right)} \right)} }}} }  \\ 
	\end{aligned},
 \label{Eq:5}
\end{equation}
where $n_{v}=n_{r}=n_{b}/2$, $s(\cdot)$ represents the similarity between two vectors; $\bm f_{y_{i}}^{t}$ is the text feature with identity $y_{i}$; $P(y_{i})$ denotes a set composed of indexes of samples with identity $y_{i}$; $|P(y_{i})|$ indicates the cardinality of $P(y_{i})$. Note that the above two formulas are employed for the training of $\mathcal{T}_{i}$, while all other components remain fixed.

After training, the learned $\mathcal{T}_{i}$ encapsulates identity-related semantic information. Regarding its associated text features $\bm f_{i}^{t}$ as the discriminative prototype, we impose constraints on $\bm f_{i}^{v}$ and $\bm f_{i}^{r}$ to achieve semantic alignment as follows:

\begin{equation}
	\begin{aligned}
L_{i2tce}^{v} =  - \frac{1}{{{n_v}}}\sum\limits_{i = 1}^{{n_v}} {{\bm q_i}\log \frac{{\exp \left( {s\left( {\bm f_i^v, \bm f_{y_{i}}^{t}} \right)} \right)}}{{\sum\nolimits_{{y_a} = 1}^{N_{c}} {\exp \left( {s\left( {\bm f_i^v, \bm f^{t}_{y_{a}}} \right)} \right)} }}}
	\end{aligned},
 \label{Eq:6}
\end{equation}

\begin{equation}
	\begin{aligned}
L_{i2tce}^{r} =  - \frac{1}{{{n_r}}}\sum\limits_{i = 1}^{{n_r}} {{\bm q_i}\log \frac{{\exp \left( {s\left( {\bm f_i^r, \bm f_{y_{i}}^{t}} \right)} \right)}}{{\sum\nolimits_{{y_a} = 1}^{N_{c}} {\exp \left( {s\left( {\bm f_i^r, \bm f^{t}_{y_{a}}} \right)} \right)} }}}
	\end{aligned},
 \label{Eq:7}
\end{equation}
where  $N_{c}$ is the total number of identities.

We jointly combine Eq. \ref{Eq:1}, Eq. \ref{Eq:2}, Eq. \ref{Eq:6}, and Eq. \ref{Eq:7} to optimize visual feature encoders $\bm E_{v}$, $\bm E_{r}$, $\bm E$, and identity classifier $\bm W$.

\subsection{Our proposed CSDN}
The above CLIP-VIReID is only our preliminary exploration of the application of CLIP in VIReID, and it still faces some challenges. This motivates us to further propose a CLIP-Driven Semantic Discovery Network (CSDN), comprising Modality-specific Prompt Learners (MsPL), Semantic Information Integration (SII), and High-level Semantic Embedding (HSE).

\subsubsection{MsPL}
Given that natural language descriptions for images of the same individual in different modalities are theoretically distinct, modality-specific prompt learners should be devised to acquire textual data. Suppose that $\mathcal{T}_{i}^{v}$: ``A photo of a [X$^{v}$]$_{1}$, [X$^{v}$]$_{2}$,\ldots, [X$^{v}$]$_{M}$, $[Cls]$ person'' and $\mathcal{T}_{i}^{r}$: ``A photo of a [X$^{r}$]$_{1}$, [X$^{r}$]$_{2}$,\ldots, [X$^{r}$]$_{M}$, $[Cls]$ person'' represent the learnable language descriptions corresponding to $\bm x_{i}$ and $\bm x_{v}$ respectively, the Eq. \ref{Eq:4} and Eq. \ref{Eq:5} should be reformulated as:
\begin{equation}
	\begin{aligned}
L_{i2t}^{1} =  &- \frac{1}{{{n_v}}}\sum\limits_{i = 1}^{{n_v}} {\log \frac{{\exp \left( {s\left( {\bm f_i^v, \bm f_{i}^{t_{v}}} \right)} \right)}}{{\sum\nolimits_{j = 1}^{{n_v}} {\exp \left( {s\left( {\bm f_i^v, \bm f_j^{{t_{v}}}} \right)} \right)} }}}  \\&- \frac{1}{{{n_r}}}\sum\limits_{i = 1}^{{n_r}} {\log \frac{{\exp \left( {s\left( {\bm f_i^r, \bm f_{i}^{{t_{r}}}} \right)} \right)}}{{\sum\nolimits_{j = 1}^{{n_r}} {\exp \left( {s\left( {\bm f_i^r, \bm f_j^{{t_{r}}}} \right)} \right)} }}} 
	\end{aligned},
 \label{Eq:8}
\end{equation}

\begin{equation}
	\begin{aligned}
	L_{t2i}^{1} =  - \frac{1}{{{n_v}}}\sum\limits_{i = 1}^{{n_v}} {\frac{1}{{\left| {P\left( {{y_i}} \right)} \right|}}\sum\limits_{{p_i} \in P\left( {{y_i}} \right)} {\log \frac{{\exp \left( {s\left( {\bm f_{{p_i}}^v, \bm f_{y_{i}}^{t_{v}}} \right)} \right)}}{{\sum\nolimits_{j = 1}^{{n_v}} {\exp \left( {s\left( {\bm f_j^v, \bm f_{y_{i}}^{t_{v}}} \right)} \right)} }}} }  \\ 
	{\rm{         }} - \frac{1}{{{n_r}}}\sum\limits_{i = 1}^{{n_r}} {\frac{1}{{\left| {P\left( {{y_i}} \right)} \right|}}\sum\limits_{{p_i} \in P\left( {{y_i}} \right)} {\log \frac{{\exp \left( {s\left( {\bm f_{{p_i}}^r, \bm f_{y_{i}}^{t_{r}}} \right)} \right)}}{{\sum\nolimits_{j = 1}^{{n_r}} {\exp \left( {s\left( {\bm f_j^r, \bm f_{y_{i}}^{t_{r}}} \right)} \right)} }}} }  \\ 
	\end{aligned},
 \label{Eq:9}
\end{equation}
where $\bm f_{i}^{t_{v}}$ and $\bm f_{i}^{t_{r}}$ represent the text features of $\mathcal{T}_{i}^{v}$ and $\mathcal{T}_{i}^{r}$.

\subsubsection{SII}
Employing $\bm f_{i}^{t_{v}}$ and $\bm f_{i}^{t_{r}}$ to facilitate the learning of $\bm f_{i}^{v}$ and $\bm f_{i}^{r}$ is feasible. However, we posit that modality-specific language descriptions may overemphasize modality-private semantic information, whereas the essence of VIReID lies in acquiring modality-shared and modality-invariant representations of pedestrians. Notably, visible and infrared images of pedestrians sharing the same identity often contain complementary information. As a result, the semantic details derived from these images also demonstrate a complementary nature. Comprehensive utilization of this information holds the potential to significantly enhance the expressive capacity of semantics. To attain this objective, we introduce the SII module.

Specifically, inspired by Non-Local \cite{Non-Local}, we design an attention fusion (AF) module to effectively integrate the semantic information present in $\bm f^{t_{v}}$ and $\bm f^{t_{r}}$. As shown in Figure \ref{Fig:2}, treating $\bm f^{t_{v}}$ as Query, and $\bm f^{t_{r}}$ as Key and Value, the integration process can be expressed as follows:
\begin{equation}
\begin{aligned}
    \bm f^{t_{s}}=\bm f^{t_{v}} + \bm W_{c} (\bm{A}\bm{W}_{v}(\bm f^{t_{r}}))
\end{aligned},
\end{equation}

\begin{equation}
\begin{aligned}
    \bm{A}=softmax\bigg( \frac{\bm W_{q}(\bm f^{t_{v}})(\bm W_{k}(\bm f^{t_{r}}))^{T}}{\sqrt{d}} \bigg)
\end{aligned},
\end{equation}
where $\bm W_{q}$, $\bm W_{k}$, $\bm W_{v}$ and $\bm W_{c}$ are four fully-connected layers; $d$ denotes the dimension of text features; $\bm f^{t_{s}}$ is the integrated text feature with rich complementary semantic information. Similarly, we deploy contrastive losses on AF to ensure the correspondence between $\bm f^{t_{s}}$ and $(\bm f^{v}, \bm f^{r})$:

\begin{equation}
	\begin{aligned}
L_{i2t}^{2} =  &- \frac{1}{{{n_v}}}\sum\limits_{i = 1}^{{n_v}} {\log \frac{{\exp \left( {s\left( {\bm f_i^v, \bm f_{i}^{t_{s}}} \right)} \right)}}{{\sum\nolimits_{j = 1}^{{n_v}} {\exp \left( {s\left( {\bm f_i^v, \bm f_j^{{t_{s}}}} \right)} \right)} }}}  \\&- \frac{1}{{{n_r}}}\sum\limits_{i = 1}^{{n_r}} {\log \frac{{\exp \left( {s\left( {\bm f_i^r, \bm f_{i}^{{t_{s}}}} \right)} \right)}}{{\sum\nolimits_{j = 1}^{{n_r}} {\exp \left( {s\left( {\bm f_i^r, \bm f_j^{{t_{s}}}} \right)} \right)} }}} 
	\end{aligned},
 \label{Eq:12}
\end{equation}

\begin{equation}
	\begin{aligned}
	L_{t2i}^{2} =  - \frac{1}{{{n_v}}}\sum\limits_{i = 1}^{{n_v}} {\frac{1}{{\left| {P\left( {{y_i}} \right)} \right|}}\sum\limits_{{p_i} \in P\left( {{y_i}} \right)} {\log \frac{{\exp \left( {s\left( {\bm f_{{p_i}}^v, \bm f_{y_{i}}^{t_{s}}} \right)} \right)}}{{\sum\nolimits_{j = 1}^{{n_v}} {\exp \left( {s\left( {\bm f_j^v, \bm f_{y_{i}}^{t_{s}}} \right)} \right)} }}} }  \\ 
	{\rm{         }} - \frac{1}{{{n_r}}}\sum\limits_{i = 1}^{{n_r}} {\frac{1}{{\left| {P\left( {{y_i}} \right)} \right|}}\sum\limits_{{p_i} \in P\left( {{y_i}} \right)} {\log \frac{{\exp \left( {s\left( {\bm f_{{p_i}}^r, \bm f_{y_{i}}^{t_{s}}} \right)} \right)}}{{\sum\nolimits_{j = 1}^{{n_r}} {\exp \left( {s\left( {\bm f_j^r, \bm f_{y_{i}}^{t_{s}}} \right)} \right)} }}} }  \\ 
	\end{aligned},
 \label{Eq:13}
\end{equation}

\subsubsection{HSE}
Our primary objective is to acquire refined visual representations for precise identity matching, leading to the proposal of the HSE module. To be specific, we employ identity loss (Eq. \ref{Eq:1}) and triplet loss (Eq. \ref{Eq:2}) to optimize $\bm f^{v}$ and $\bm f^{r}$, ensuring their discriminative. Additionally, with the guidance of $\bm f^{t_{s}}$, we employ the following losses to ensure their semantic richness, and, consequently, promote their modality invariance:
\begin{equation}
	\begin{aligned}
L_{i2tce}^{v_{s}} =  - \frac{1}{{{n_v}}}\sum\limits_{i = 1}^{{n_v}} {{\bm q_i}\log \frac{{\exp \left( {s\left( {\bm f_i^v, \bm f_{y_{i}}^{t_{s}}} \right)} \right)}}{{\sum\nolimits_{{y_a} = 1}^{N_{c}} {\exp \left( {s\left( {\bm f_i^v, \bm f^{t_{s}}_{y_{a}}} \right)} \right)} }}}
	\end{aligned},
 \label{Eq:14}
\end{equation}

\begin{equation}
	\begin{aligned}
L_{i2tce}^{r_{s}} =  - \frac{1}{{{n_r}}}\sum\limits_{i = 1}^{{n_r}} {{\bm q_i}\log \frac{{\exp \left( {s\left( {\bm f_i^r, \bm f_{y_{i}}^{t_{s}}} \right)} \right)}}{{\sum\nolimits_{{y_a} = 1}^{N_{c}} {\exp \left( {s\left( {\bm f_i^r, \bm f^{t_{s}}_{y_{a}}} \right)} \right)} }}}
	\end{aligned},
 \label{Eq:15}
\end{equation}

Note that this module is only designed to train visual encoders and the identity classifier, and the total loss can be expressed as follows:

\begin{equation}
L_{total}=L_{id}+\lambda_{1}L_{wrt}+\lambda_{2}L_{i2tce}^{v_{s}}+\lambda_{3}L_{i2tce}^{r_{s}},
\label{Eq:16}
\end{equation}
where $\lambda_{1}$, $\lambda_{2}$, and $\lambda_{3}$ are hyper-parameters that balance the contribution of each loss term.

\subsection{Training and Inference}

The proposed method delineates the entire training process into three stages. Initially, in the first stage, with the constraints of Eq. \ref{Eq:8} and Eq. \ref{Eq:9}, we train the MsPL equipped with two learnable prompts, responsible for generating textual descriptions ($\mathcal{T}^{v}$ and $\mathcal{T}^{r}$) corresponding to both visible and infrared images. Moving to the second stage, we impose constraints as defined in Eq. \ref{Eq:12} and Eq. \ref{Eq:13} to train the SII that comprises attention fusion ($\bm W_{q}$, $\bm W_{k}$, $\bm W_{v}$ and $\bm W_{c}$ ), tasked with integrating the high-level semantic information from text descriptions generated in the initial stage, thereby attaining more comprehensive semantics. Finally, the third stage involves training visual encoders ($\bm E_{v}$, $\bm E_{r}$, and $\bm E$) and identity classifier ($\bm W$) with the constraint specified in Eq. \ref{Eq:16}, fostering discriminative and modality-invariant learning of visual representations. Notably, during the inference phase, only the features extracted by the visual encoder are utilized to measure similarity using cosine distance for identity matching. The remaining components are not required, ensuring that the practicality of our proposed CSDN remains uncompromised. Algorithm \ref{Alg:1} provides a comprehensible introduction to the entire training process.

\begin{algorithm}[!t]\small
 \caption{CLIP-Driven Semantic Discovery Network for Visible-Infrared Person Re-Identification.}\label{Alg:1}
\begin{algorithmic}
\STATE {\textbf{Input:} Sample set $\bm D=\{\bm x_{i}^{v}, \bm x_{i}^{r}, y_{i}\}_{i=1}^{N}$, pre-trained CLIP, hyper-parameters $\lambda_{1}$, $\lambda_{2}$ and $\lambda_{3}$, number of iterations $T_1$, $T_2$, and $T_3$.\\}
\STATE {\textbf{Output:} The trained visual encoder $\bm E_{v}$, $\bm E_{r}$, and $\bm E$.\\
\begin{flushleft}
~1:Initialize $\mathcal{T}^{v}$, $\mathcal{T}^{r}$, $\bm W_{q}$, $\bm W_{k}$, $\bm W_{v}$, $\bm W_{c}$, $\bm E_{v}$, $\bm E_{r}$ $\bm E$, and $\bm W$.\\
~~~\textbf{Stage I}: Training the MsLP\\
~2:\textbf{for} \emph{iter}=1, $\cdots$, $T_{1}$ \textbf{do}\\
~3:\qquad Update $\mathcal{T}^{v}$ and $\mathcal{T}^{r}$ by minimizing Eq. \ref{Eq:8} and Eq. \ref{Eq:9}.\\
~4:\textbf{end for}\\
~~~\textbf{Step \uppercase\expandafter{\romannumeral2}}: Training the SII\\
~5:\textbf{for} \emph{iter}=1, $\cdots$, $T_{2}$ \textbf{do}\\
~6:\qquad Update $\bm W_{q}$, $\bm W_{k}$, $\bm W_{v}$, and $\bm W_{c}$ by minimizing Eq. \ref{Eq:12} and \\~~~~~~~~ Eq. \ref{Eq:13}.\\
~7:\textbf{end for}\\
~~~\textbf{Step \uppercase\expandafter{\romannumeral3}}: Training the HSE\\
~8:\textbf{for} \emph{iter}=1, $\cdots$, $T_{3}$ \textbf{do}\\
~9:\qquad Update $\bm E_{v}$, $\bm E_{r}$, $\bm E$ and $\bm W$ by minimizing Eq. \ref{Eq:16}.\\
10:\textbf{end for}\\
\end{flushleft}}
\end{algorithmic}
\end{algorithm}

\section{Experiments}

\subsection{Experimental Settings}

\subsubsection{Datasets} 
We conduct experiments on two public VI-ReID datasets, namely SYSU-MM01 \cite{VIREID} and RegDB \cite{RegDB}.

\textbf{SYSU-MM01} is currently the most challenging large-scale dataset specially constructed for cross-modality person ReID, encompassing 287,628 visible images and 15,792 infrared images of 491 identities. Following the standard protocol \cite{VIREID}, the training set comprises 22,258 visible images and 11,909 infrared images of 395 pedestrians, and the testing set includes images of 96 pedestrians. All images were captured from 4 visible cameras and 2 infrared cameras positioned both indoors and outdoors. In the testing phase, infrared images serve as the query and visible ones constitute the gallery. The dataset offers two testing modes: all-search, involving all visible images from indoors and outdoors, and indoor-search, with only images collected from two indoor cameras contributing to the test. Additionally, the dataset provides single-shot and multi-shot gallery settings, indicating the selection of 1 or 10 visible images per identity as the gallery, respectively. This study comprehensively assesses the performance across all these settings.

\begin{table*}[!ht]\scriptsize
\centering {\caption{Performance comparison with state-of-the-art methods on SYSU-MM01. The upper section enumerates generative-based methods, with the optimum performance indicated by '\underline{*}'. The lower section catalogs non-generative-based methods, with the optimal performance marked by '\uwave{*}'. The performance of our method is highlighted in bold. '-' denotes that no reported result is available.}\label{Tab:1}
\renewcommand\arraystretch{1.2}
\begin{tabular}{c|c|cccc|cccc|cccc|cccc}
\hline
 \hline
  \multirow{3}*{Methods} & \multirow{3}*{Venue} & \multicolumn{8}{c|}{All-Search} & \multicolumn{8}{c}{Indoor-Search}\\
  
  \cline{3-18} & & \multicolumn{4}{c|}{Single-Shot} & \multicolumn{4}{c|}{Multi-Shot} & \multicolumn{4}{c|}{Single-Shot} & \multicolumn{4}{c}{Multi-Shot}\\
  
  & & R1 & R10 & R20 & mAP & R1 & R10 & R20 & mAP & R1 & R10 & R20 & mAP & R1 & R10 & R20 & mAP\\
\hline

  D$^{2}$RL \cite{D2RL} & CVPR'19 & 28.9 & 70.6 & 82.4 & 29.2 & - & - & - & - & - & - & - & - & - & - & - & - \\

  AlignGAN \cite{AlignGAN} & ICCV'19 & 42.4 & 85.0 & 93.7 & 40.7 & 51.5 & 89.4 & 95.7 & 33.9 & 45.9 & 87.6 & 94.4 & 54.3 & 57.1 & 92.7 & 97.4 & 45.3 \\

  Hi-CMD \cite{HiCMD} & CVPR'20 & 34.9 & 77.5 & - & 35.9 & - & - & - & -& - & - & - & - & - & - & - & -\\

  JSIA \cite{JSIA} & AAAI'20 & 38.1 & 80.7 & 89.9 & 36.9 & 45.1 & 85.7 & 93.8 & 29.5 & 43.8 & 86.2 & 94.2 & 52.9 & 52.7 & 91.1 & 96.4 & 42.7 \\

  X-Modality \cite{X-modality} & AAAI'20 & 49.9 & 89.7 & 95.9 & 50.7 & - & - & - & - & - & - & - & - & - & - & - & -\\


  CECNet \cite{CECNet} & TCSVT'22 & 53.3 & 89.8 & 95.6 & 51.8 & - & - & - & - & 60.6 & 94.2 & 98.1 & 62.8 & - & - & - & - \\

  RBDF \cite{RBDF} & TCYB'22 & 57.6 & 85.8 & 91.2 & 54.4 & - & - & - & - & - & - & - & - & - & - & - & - \\

  TSME \cite{TSME} & TCSVT'22 & \underline{64.2} & \underline{95.1} & \underline{98.7} & \underline{61.2} & \underline{70.3} & \underline{96.7} & \underline{99.2} & \underline{54.3} & \underline{64.8} & \underline{96.9} & \underline{99.3} & \underline{71.3} & \underline{76.8} & \underline{98.8} & \underline{99.8} & \underline{65.0} \\

  \hline

  Zero-Pad \cite{VIREID} & ICCV'17 & 14.8 & 54.1 & 71.3 & 15.9 & 19.1 & 61.4 & 78.4 & 10.8 & 20.5 & 68.8 & 85.7 & 26.9 & 24.4 & 75.8 & 91.3 & 18.6 \\


  MSR \cite{MSR} & TIP'19 & 37.3 & 83.4 & 93.3 & 38.1 & 43.8 & 86.9 & 95.6 & 30.4 & 39.6 & 89.2 & 97.6 & 50.8 & 46.5 & 93.5 & 98.8 & 40.0 \\

  DFE \cite{DFE} & MM'19 & 48.7 & 88.8 & 95.2 & 48.5 & 54.6 & 91.6 & 96.8 & 42.1 & 52.2 & 89.8 & 95.8 & 59.6 & 59.6 & 94.4 & 98.0 & 50.6 \\

  FMSP \cite{FMSP} & IJCV'20 & 43.5 & - & - & 44.9 & - & - & - & - & 48.6 & - & - & 57.5 & - & - & - & -\\

  DDAG \cite{DDAG} & ECCV'20 & 54.7 & 90.3 & 95.8 & 53.0 & - & - & - & - & 61.0 & 94.0 & 98.4 & 67.9 & - & - & - & -\\

  AGW \cite{AGW} & TPAMI'21 & 47.5 & 84.3 & 92.1 & 47.6 & - & - & - & - & 54.1 & 91.1 & 95.9 & 62.9 & - & - & - & - \\

  LbA \cite{LbA} & ICCV'21 & 55.4 & - & - & 54.1 & - & - & - & - & 58.4 & - & - & 66.3 & - & - & - & - \\

  NFS \cite{NFS} & CVPR'21 & 56.9 & 91.3 & 96.5 & 55.4 & 63.5 & 94.4 & 97.8 & 48.5 & 62.7 & 96.3 & 99.0 & 69.7 & 70.0 & 97.7 & 99.5 & 61.4 \\


  MSO \cite{MSO} & MM'21 & 58.7 & 92.0 & 97.2 & 56.4 & 65.8 & 94.3 & 98.2 & 49.5 & 63.0 & 96.6 & 99.0 & 70.3 & 72.0 & 97.7 & 99.6 & 61.6 \\

  MPANet \cite{MPANet} & CVPR'21 & 70.5 & 96.2 & 98.8 & 68.2 & \uwave{75.5} & \uwave{97.7} & \uwave{99.4} & \uwave{62.9} & 76.7 & 98.2 & 99.5 & 80.9 & \uwave{84.2} & \uwave{99.6} & \uwave{99.9} & \uwave{75.1} \\

  CAJ \cite{CAJ} & ICCV'21 & 69.8 & 95.7 & 98.4 & 66.8 & - & - & - & - & 76.2 & 97.8 & 99.4 & 80.3 & - & - & - & - \\
  
  PIC \cite{PIC} & TIP'22 & 57.5 & 89.3 & - & 55.1 & - & - & - & - & 60.4 & - & - & 67.7 & - & - & - & - \\

   DML \cite{DML} & TCSVT'22 & 58.4 & 91.2 & 96.9 & 56.1 & 62.2 & 93.4 & 97.8 & 49.6 & 62.4 & 95.2 & 98.7 & 69.5 & 66.4 & 96.7 & 99.5 & 60.0 \\

  SPOT \cite{SPOT} & TIP'22 & 65.3 & 92.7 & 97.0 & 62.2 & - & - & - & - & 69.4 & 96.2 & 99.1 & 74.6 & - & - & - & - \\


   DCLNet \cite{DCLNet} & MM'22 & 70.8 & - & - & 65.3 & - & - & - & - & 73.5 & - & - & 76.8 & - & - & - & - \\



  DSCNet \cite{DSCNet} & TIFS'22 & \uwave{73.8} & \uwave{96.2} & \uwave{98.8} & \uwave{69.4} & - & - & - & - & \uwave{79.3} & \uwave{98.3} & \uwave{99.7} & \uwave{82.6} & - & - & - & - \\

  GUR \cite{GUR} & ICCV'23 & 63.5 & - & - & 61.6 & - & - & - & - & 71.1 & - & - & 76.2 & - & - & - & - \\

  CMTR \cite{CMTR} & TMM'23 & 65.4 & 94.4 & 98.1 & 62.9 & 71.9 & 96.3 & 99.0 & 57.0 & 71.4 & 97.1 & 99.2 & 76.6 & 80.0 & 98.5 & 99.7 & 69.4 \\

  PMT \cite{PMT} & AAAI'23 & 67.5 & 95.3 & 98.6 & 64.9 & - & - & - & - & 71.6 & 96.7 & 99.2 & 76.5 & - & - & - & - \\


  CAJ$_{+}$ \cite{CAJ+} & TPAMI'23 & 71.4 & \uwave{96.2} & 98.7 & 68.1 & - & - & - & - & 78.3 & \uwave{98.3} & \uwave{99.7} & 81.9 & - & - & - & - \\



  \hline

  Ours & - & \bf{75.2} & \bf{96.6} & \bf{98.8} & \bf{71.8} & \bf{80.6} & \bf{98.3} & \bf{99.7} & \bf{66.3} & \bf{82.0} & \bf{98.7} & \bf{99.5} & \bf{85.0} & \bf{88.5} & \bf{99.6} & \bf{99.9} & \bf{80.4} \\

  \hline\hline
\end{tabular}}
\end{table*}

\textbf{RegDB} is a small-scale VI-ReID dataset comprising 8,240 images featuring 412 pedestrians. Each identity has 10 visible images and 10 infrared images, all collected from a single camera for both modalities. Adhering to the evaluation protocol \cite{AlignGAN}, we randomly select 2,060 visible images and 2,060 infrared images of 206 pedestrians as the training set, and the remaining ones constitute the testing set. This dataset provides two evaluation scenarios: visible-to-infrared and infrared-to-visible retrievals. It is noteworthy that the dataset underwent random division into training and testing sets 10 times for experiments, and the average results were considered for the final performance, ensuring accuracy and fairness in the evaluation.

\subsubsection{Evaluation Metrics} We assess the performance with the general evaluation metrics: Cumulative Matching Characteristics (CMC) and mean Average Precision (mAP). For the CMC, Rank-1, Rank-10, and Rank-20 are reported.

\subsubsection{Implementation Details} We implement the proposed CSDN with the Pytorch deep learning framework, and all experiments are deployed on one GTX3090 GPU. We adopt CLIP \cite{CLIP} as the backbone, employing two parallel shallow layers for modality-specific feature extraction and the remaining four shared deep convolutional blocks for modality-shared features. Input images are uniformly resized to 288$\times$144, and common data augmentation strategies such as random flipping, padding, and cropping are applied. In the first and second stages, we train two modality-specific prompt learners and attention fusion modules, each for 60 epochs. The initial learning rate is set to $3\times10^{-4}$ and decayed following a cosine schedule \cite{CLIP-ReID}. Subsequently, in the third stage, we exclusively train the visual encoder and classifier for 120 epochs. The learning rate in this stage undergoes a linear increase from $3\times10^{-6}$ to $3\times10^{-4}$ in the first 10 epochs and decayed by 0.1 at the 40th epoch and the 70th epoch. The batch size is set to 64 with 8 pedestrians, each pedestrian has 4 visible images and 4 infrared images. The training process of the proposed CSDN employs the Adam optimizer \cite{Adam}. The hyper-parameters are set to $\lambda_{1}=0.15$, $\lambda_{2}=0.05$ and $\lambda_{3}=0.1$, respectively.

\subsection{Comparison with State-of-the-art Methods}
In this section, we compare the proposed CSDN with state-of-the-art methods on two widely used VI-ReID benchmarks. The results are summarized in Table\ref{Tab:1} and Table\ref{Tab:2}.

\subsubsection{SYSU-MM01} We first conduct experiments on SYSU-MM01. As illustrated in Table \ref{Tab:1}, the proposed CSDN consistently outperforms state-of-the-art methods across all settings. Specifically, under all-search testing and single-shot gallery modes, our algorithm, without a requirement for a generation process, achieves 75.2\% in Rank-1 recognition rate and 71.8\% in mAP accuracy, surpassing TSME \cite{TSME} by 11.0\% and 10.6\%, respectively. Notably, it also exhibits a substantial performance advantage over TSME under the indoor-search testing mode. In comparison to DSCNet \cite{DSCNet}, a superior non-generative-based method with 73.8\% (69.4\%) and 79.3\% (82.6) Rank-1 (mAP) accuracy under all search (single-shot) and indoor-search (single-shot) testing modes, our CSDN demonstrates a significant improvement of 1.4\% (2.7\%) in Rank-1 accuracy and 2.4\% (2.4\%) in mAP. Furthermore, MPANet \cite{MPANet} obtains 75.5\% (84.2\%) Rank-1 accuracy and 62.9\% (75.1\%) mAP recognition rates under all-search (multi-shot) and indoor-search (multi-shot) testing modes. In contrast, our proposed CSDN improves the Rank-1 and mAP by 5.1\% (4.3\%) and 3.4\% (5.3\%).

\subsubsection{RegDB} To further evaluate the efficacy of our approach, we conduct experiments on the RegDB dataset and compare it with previous studies, as delineated in Table \ref{Tab:2}. It can be seen that generative-based methods have gained satisfactory performance, exemplified by TSME \cite{TSME}, which boasts 87.3\% (86.4\%) Rank-1 accuracy and 76.9\% (75.7\%) mAP under the visible-to-infrared (infrared-to-visible) searching mode. This substantial improvement is primarily due to the expanded scale of RegDB through generative augmentation. Despite this, our CSDN surpasses it by 1.7\% (1.8\%) and 7.8\% (7.1\%) in Rank-1 accuracy and mAP under the visible-to-infrared (infrared-to-visible) testing mode. Additionally, some recent non-generative-based methods demonstrate superior performance, such as CMTR \cite{CMTR} and CAJ$_{+}$ \cite{CAJ+}. In comparison, our CSDN outperforms CMTR by 3.1\% (2.1\%) in mAP and surpasses CAJ$_{+}$ by 3.4\% (3.4\%) in Rank-1 accuracy under the visible-to-infrared (infrared-to-visible) testing mode,  exhibiting a considerable advantage.

The results above comprehensively demonstrate the superiority of our method across two benchmarks. This excellence is primarily attributable to key factors: (1) We introduce CLIP to facilitate VIReID learning, which guides the model to acquire modality-invariant high-level semantic information, such as gender, hairstyle, clothing, etc. This alleviates the challenge of aligning visible and infrared visual features. (2) Instead of simply leveraging CLIP's pre-trained model, we adopt the paradigm of learnable prompts to generate language descriptions conducive to visual feature learning, enabling us to fully exploit the powerful capability of CLIP. (3) Considering the divergent emphasis on language descriptions corresponding to images of different modalities, we design modality-specific learnable prompts to generate respective semantic information for visible and infrared images. Building on this, recognizing the complementary nature of semantics across two modalities, we devise an attention fusion module. This module yields a more comprehensive text feature, further facilitating the alignment of visual representations across different modalities.

\begin{table}[!ht]\scriptsize
\centering {\caption{Performance comparison with state-of-the-art methods on RegDB.}\label{Tab:2}
\renewcommand\arraystretch{1.2}
\begin{tabular}{c|cccc|cccc}
\hline
 \hline
  \multirow{2}*{Methods} & \multicolumn{4}{c|}{Visible to Infrared} & \multicolumn{4}{c}{Infrared to Visible} \\
  \cline{2-9} & R1 & R10 & R20 & mAP & R1 & R10 & R20 & mAP\\
  \hline

  D$^{2}$RL \cite{D2RL} & 43.4 & 66.1 & 76.3 & 44.1 & - & - & - & - \\

  AlignGAN \cite{AlignGAN} & 57.9 & - & - & 53.6 & 56.3 & - & - & 53.4 \\

  Hi-CMD \cite{HiCMD} & 70.9 & 86.3 & - & 66.0 & - & - & - & - \\

  JSIA \cite{JSIA} & 48.5 & - & - & 49.3 & 48.1 & - & - & 48.9 \\

  X-Modality \cite{X-modality} & - & - & - & - & 62.2 & 83.1 & 91.7 & 60.1 \\

  CECNet \cite{CECNet} & 82.3 & 92.7 & 95.4 & 78.4 & 78.9 & 91.9 & 95.4 & 75.5 \\

  RBDF \cite{RBDF} & 79.8 & 93.5 & 96.9 & 76.7 & 76.2 & 90.7 & 94.8 & 73.9 \\

  TSME \cite{TSME} & \underline{87.3} & \underline{97.1} & \underline{98.9} & \underline{76.9} & \underline{86.4} & \underline{96.3} & \underline{98.2} & \underline{75.7} \\

  \hline

  Zero-Pad \cite{VIREID} & 17.8 & - & - & 18.9 & 16.7 & - & - & 17.9 \\

  MSR \cite{MSR} & 48.4 & 70.3 & 79.9 & 48.6 & - & - & - & - \\

  DFE \cite{DFE} & 70.1 & 86.3 & 91.9 & 69.1 & 67.9 & 85.5 & 91.4 & 66.7 \\

  FMSP \cite{FMSP} & 65.0 & 83.7 & - & 64.5 & - & - & - & - \\

   DDAG \cite{DDAG} & 69.3 & 86.1 & 91.4 & 63.4 & 68.0 & 85.1 & 90.3 & 61.8 \\

   AGW \cite{AGW} & 70.0 & - & - & 66.3 & 70.4 & - & - & 65.9 \\

  LbA \cite{LbA} & 74.1 & - & - & 67.6 & 72.4 & - & - & 65.4 \\ 

  NFS \cite{NFS} & 80.5 & 91.9 & 95.0 & 72.1 & 77.9 & 90.4 & 93.6 & 69.7 \\


  MSO \cite{MSO} & 73.6 & 88.6 & - & 66.9 & 74.6 & 88.7 & - & 67.5 \\

  MPANet \cite{MPANet} & 83.7 & - & - & 80.9 & 82.8 & - & - & 80.7 \\

   CAJ \cite{CAJ} & 85.0 & \uwave{95.4} & \uwave{97.5} & 79.1 & 84.7 & 95.3 & 97.5 & 77.8 \\

  PIC \cite{PIC} & 83.6 & - & - & 79.6 & 79.5 & - & - & 77.4 \\

  DML \cite{DML} & 84.3 & - & - & 77.6 & 83.6 & - & - & 77.0 \\

  SPOT \cite{SPOT} & 80.3 & 93.4 & 96.4 & 72.4 & 79.3 & 92.7 & 96.0 & 72.2 \\

  DCLNet \cite{DCLNet} & 81.2 & - & - & 74.3 & 78.0 & - & - & 70.6 \\

  DSCNet \cite{DSCNet} & 85.3 & - & - & 77.3 & 83.5 & - & - & 75.1 \\

  GUR \cite{GUR} & 73.9 & - & - & 70.2 & 75.0 & - & - & 56.2 \\

  CMTR \cite{CMTR} & \uwave{88.1} & - & - & \uwave{81.6} & \uwave{84.9} & - & - & \uwave{80.7} \\

  PMT \cite{PMT} & 84.8 & - & - & 76.5 & 84.1 & - & - & 75.1 \\

  CAJ$_{+}$ \cite{CAJ+} & 85.6 & \uwave{95.4} & \uwave{97.5} & 79.7 & 84.8 & \uwave{95.8} & \uwave{97.7} & 78.5 \\



  \hline

  Ours & \bf{89.0} & \bf{96.1} & \bf{97.9} & \bf{84.7} & \bf{88.2} & \bf{95.1} & \bf{96.6}& \bf{82.8} \\

  \hline\hline
\end{tabular}}
\end{table}

\subsection{Ablation Studies}

In this section, we undertake ablation studies to validate the efficacy of each module integrated into the proposed method. To be specific, we first qualitatively analyze the performance gain associated with each component, and the results are summarized in Table \ref{Tab:3}. The Baseline denotes the general model commonly employed in extant studies, yielding a Rank-1 accuracy of 70.4\% and mAP of 66.8\%. Furthermore, we perform a quantitative assessment of the model's effectiveness through class activation maps (CAMs) \cite{CAM} that highlight spatially discriminative regions within the image. All experiments are conducted under indoor-search testing and multi-gallery mode.

\begin{table}[!ht]\small
\centering {\caption{Ablation studies of the proposed CSDN.}\label{Tab:3}
\renewcommand\arraystretch{1.2}
\begin{tabular}{c|c|cccc|cc}
\hline
 \hline
  \multicolumn{2}{c|}{Methods} & PL & MsPL & SII & HSE & R1 & mAP \\
\hline

  \multicolumn{2}{c|}{Baseline} &  &  &  & & 70.4 & 66.8 \\

  \multicolumn{2}{c|}{CLIP Pre-trained} &  &  &  & & 71.5 & 67.2 \\

  \hline

  \multicolumn{2}{c|}{CLIP-VIReID} & \checkmark &  &  & \checkmark & 73.9 & 71.7 \\

  \hline

  \multirow{3}*{CSDN} & \#1 &  & \checkmark &  & \checkmark & 73.5 & 70.9
  \\

  & \#2 & & \checkmark & \checkmark & \checkmark & 73.7 & 71.0 \\

  & \#3 & & \checkmark & \checkmark & \checkmark & \bf{75.2} & \bf{71.8} \\

  \hline\hline
\end{tabular}}
\end{table}

\subsubsection{Effectiveness of CLIP Pre-trained model}
This study is committed to acquiring semantically rich visual representations to facilitate VIReID learning. Recognizing that CLIP possesses the ability to sense high-level semantics related to target pedestrians, we substitute the backbone (ResNet) of the Baseline with the visual encoder from the CLIP's pre-trained model. It can be seen that the Rank-1 accuracy and mAP are improved by 1.1\% and 0.4\%, respectively, which substantiates the rationality of our motivation and affirms the effectiveness of the technology.

\subsubsection{Effectiveness of CLIP-VIReID}
As we mentioned previously, a mere adaptation of CLIP's pre-trained model to VIReID falls short of fully harnessing CLIP's potent capability, as evidenced by the limited performance improvement indicated in the above experimental results. In light of this limitation, similar to CLIP-ReID, we employ the prompt learner (PL) paradigm to generate language descriptions for pedestrian images. In particular, these descriptions may encompass high-level semantic information such as gender, hairstyle, and clothing, unaffected by different modalities, thus potentially promoting the alignment of visible and infrared features. As illustrated in Table \ref{Tab:3}, CLIP-VIReID further improves the Rank-1 from 71.5\% to 73.9\% and mAP from 67.2\% to 71.7\%, validating its effectiveness.

\begin{figure}[th!]
\centering
{\includegraphics[height=3.8in,width=3.5in,angle=0]{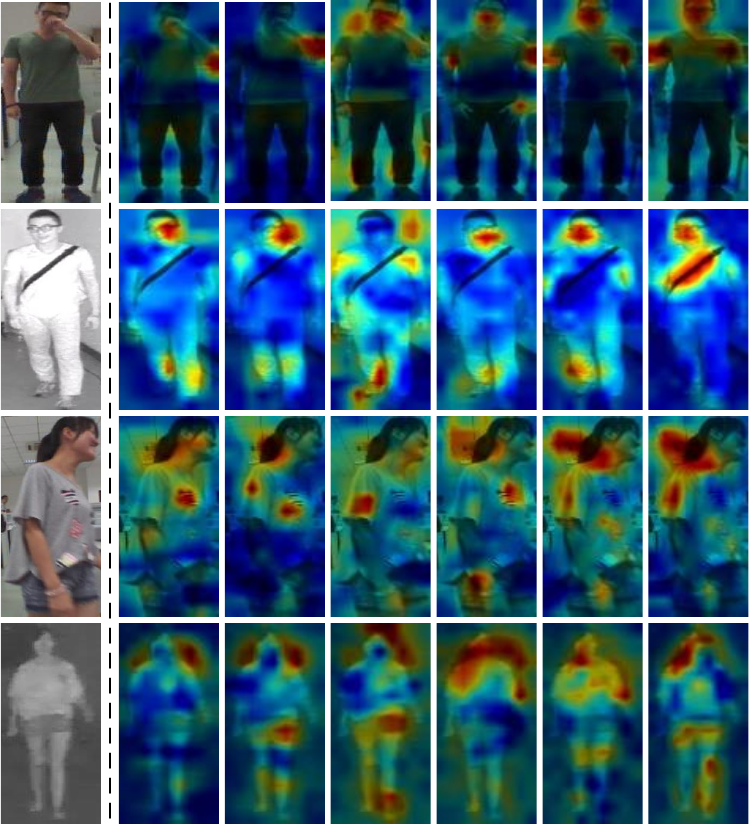}}
\caption{Visualization of spatial discriminative regions. The images are arranged from left to right in the following order: original image, heatmap obtained by Baseline, CLIP Pre-trained, CLIP-VIReID, CSDN\#1, CSDN\#2, and CSDN\#3.}
\label{Fig:4}
\end{figure}

\subsubsection{Effectiveness of MsPL}
In CLIP-VIReID, we introduce a modality-shared prompt learner for pedestrians who share the same identity across diverse modalities. Nevertheless, distinctions arise in the descriptions corresponding to images of different modalities. For instance, descriptions of visible images may highlight the color of the pedestrian's clothing, whereas those of infrared images may emphasize the pedestrian's body shape. Consequently, we devise two modality-specific prompt learners tailored to describe pedestrian images in each modality. Results (CSDN\#1) presented in Table \ref{Tab:3} demonstrate the superior effectiveness of this approach over using solely the CLIP pre-training model, yielding improvements of 2.0\% and 3.7\% in Rank-1 and mAP. It is essential to note that while its performance is commendable, it does not surpass that of CLIP-VIReID. We speculate that this outcome may be attributed to the modality-private text features overlooking shared information, increasing the difficulty of modality alignment. This observation prompts our consideration to further fully leverage modality-private language descriptions to guide effective model learning.

\subsubsection{Effectiveness of SII}
The language descriptions corresponding to images from different modalities exhibit both differences and complementarity. Their combination enables a more comprehensive portrayal of pedestrian characteristics. To this end, we introduce an attention fusion module designed to amalgamate modality-private text features. The resultant integrated semantic information serves to further bridge the correlation between the visual representations of visible and infrared images. As evident in Table \ref{Tab:3}, the recognition performance of CSDN\#2 and CSDN\#3 improves compared to CSDN\#1, compensating for shared information overlooked by modality-specific language descriptions. Additionally, it is noteworthy that CSDN\#2, treating infrared text features as Key and visible ones as Query and Value, exhibits marginal performance improvement and does not surpass CLIP-VIReID. We attribute this observation to the limited information in the text description of infrared images, suggesting that utilizing infrared text features as Key may not be the optimal solution for integrating semantic information.

\subsubsection{Effectiveness of the proposed CSDN}
In the context of the preceding discussion and the experimental verification presented above, our proposed CSDN incorporates two modality-specific prompt learners dedicated to visible and infrared images, respectively. Building on this basis, the designed attention fusion module designates the text features corresponding to visible images as Key and those corresponding to infrared images as Query and Value. This strategic arrangement aims to acquire semantically rich text features, thereby guiding the learning of the VIReID model—referred to as CSDN\#3 in Table \ref{Tab:3}. Notably, this algorithm attains optimal performance, boasting a Rank-1 accuracy of 75.2\% and an mAP of 71.8\%. Moreover, the visualization in Fig.\ref{Fig:4} vividly illustrates that the proposed CSDN enables the model to focus on more salient regions of pedestrians. These results demonstrate the positive contributions of each module to the proposed method. In summary, our CSDN is both rational and effective, exhibiting superior performance.

\subsection{Parameter Analysis}
\begin{figure*}[t!]
  \centering
  
  \subfigbottomskip=-1pt
  \subfigcapskip=-1pt
  \subfigure[$\lambda_{1}$] {\includegraphics[height=1.8in,width=2.4in,angle=0]{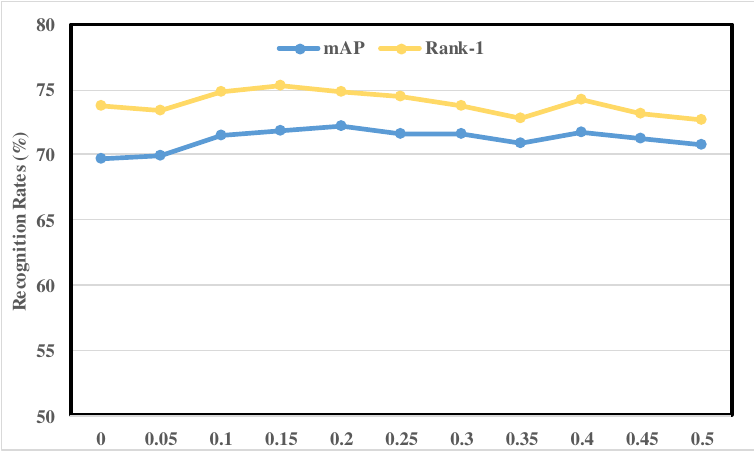}}
  \subfigure[$\lambda_{2}$] {\includegraphics[height=1.8in,width=2.4in,angle=0]{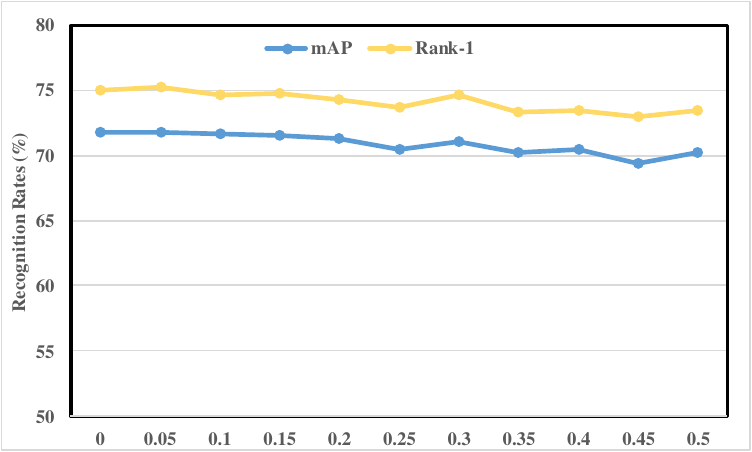}}
  \subfigure[$\lambda_{3}$] {\includegraphics[height=1.8in,width=2.4in,angle=0]{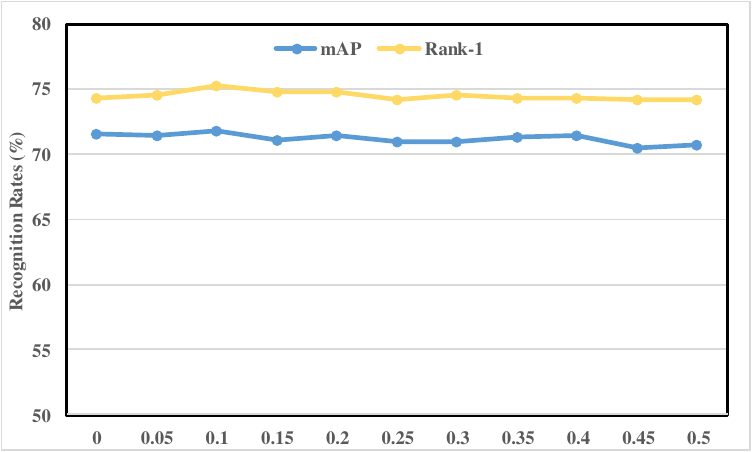}}\\
  \centering\caption{The effect analysis on different hyper-parameters $\lambda_{1}$, $\lambda_{2}$, and $\lambda_{3}$. Rank-1 accuracy and mAP are reported. Note that when one of the hyper-parameters is analyzed, the remaining two are fixed at the optimal values.}
  \label{Fig:5}
  \vspace{-0.3cm}
\end{figure*}

In the proposed CSDN, three hyper-parameters denoted as $\lambda_{1}$, $\lambda_{2}$, and $\lambda_{3}$, are employed to control the relative importance of distinct loss terms throughout the entire training process. This section encompasses a parameter analysis to ascertain the optimal value for each hyper-parameter. The corresponding results are presented in Figure \ref{Fig:5}.

\subsubsection{Effect of $\lambda_{1}$ on Performance}
In line with prevalent practices in existing VIReID studies, our proposed CSDN incorporates the weighted regularized triplet loss to minimize the feature distance among identical pedestrians while maximizing the distance between features of distinct pedestrians. We introduce the hyper-parameter $\lambda_{1}$ to regulate the weighting of this loss term. In Figure \ref{Fig:5}, it is evident that the Rank-1 accuracy attains its zenith at $\lambda_{1}=0.15$, affirming 0.15 as the optimal value for $\lambda_{1}$.

\subsubsection{Effect of $\lambda_{2}$ on Performance}
In the proposed CSDN, we expect to leverage integrated text features to guide the learning of visual representations. To realize this objective, we introduce the loss term $L_{i2t}^{3,v}$ to facilitate the learning process of visible visual representations. The hyper-parameter $\lambda_{2}$ is designated to control its relative significance. As depicted in Figure \ref{Fig:5}, our CSDN attains its peak recognition performance at $\lambda_{2}=0.05$, signifying that 0.05 is the identified optimal value for $\lambda_{2}$. In addition, when $\lambda_{2}=0.05$ compared to $\lambda_{2}=0$, the model exhibits improved Rank-1 accuracy and mAP, demonstrating the effectiveness of the loss $L_{i2t}^{3,v}$.

\subsubsection{Effect of $\lambda_{3}$ on Performance}
The hyper-parameter $\lambda_{3}$ plays a role in balancing the contribution of the loss term $L_{i2t}^{3,r}$, specifically designed to guide the learning of infrared visual features. Experimental results shown in Figure \ref{Fig:5} indicate that the optimal value for the hyper-parameter $\lambda_{3}$ is 0.1, aligning with the highest recognition rate. In addition, setting $\lambda_{3}$ to 0.05, similar to $\lambda_{2}$, leads to a performance decline. This is attributed that infrared visual features contain less information than visible ones, warranting a more stringent constraint.

\begin{table}[!ht]\small
\centering {\caption{Model complexity analysis. 'TP': The number of trained parameters. ‘IT’: The inference time. 'S' represents the stage and 'To' denotes the total number of trainable parameters.}\label{Tab:4}
\renewcommand\arraystretch{1.2}
\begin{tabular}{c|c|ccc|c|c}
\hline
 \hline
  \multicolumn{2}{c|}{\multirow{2}*{Methods}} & \multicolumn{4}{c|}{TP} & \multirow{2}*{IT} \\

  \cline{3-6} \multicolumn{2}{c|}{} & S1 & S2 & S3 & To &\\

\hline

  \multicolumn{2}{c|}{Baseline} & - & - & - & 24.3M & 52s \\

  \multicolumn{2}{c|}{CLIP Pre-trained} & - & - & - & 24.4M & 52s \\

  \hline

  \multicolumn{2}{c|}{CLIP-VIReID} & 0.8M & 39.8M & - & 40.6M & 59s \\

  \hline

  \multirow{3}*{CSDN} & \#1 & 1.6M & - & 39.8M & 41.4M & 59s \\

  & \#2 & 1.6M & 4.2M & 39.8M & 45.6M & 59s \\

  & \#3 & 1.6M & 4.2M & 39.8M & 45.6M & 59s \\

  \hline\hline
\end{tabular}}
\end{table}

\subsection{Further Discussion}

\subsubsection{Model Complexity}
In this study, our core motivation is to learn visual representation with rich semantic information to alleviate the challenge of modality alignment in VIReID. Fortunately, CLIP, renowned for its capacity to establish connections between visual representations and text features, serves as our foundational framework. It is crucial to emphasize that our adoption of CLIP is not driven by its sheer size, and the performance improvement of our method does not rely on this but on the guidance of natural language descriptions. In this section, we meticulously tabulated the model parameters, measured the model inference time, and presented the results in Table \ref{Tab:4}. Our findings and conclusions are delineated below: (1) The visual encoder backbone of CLIP encompasses two types: CNN and Transformer \cite{Transformer}. In the proposed CSDN, we opt for the former as our foundational framework. As evident from Table \ref{Tab:4}, the number of total trained parameters of CLIP Pre-trained align closely with the Baseline. This substantiates the argument above that the performance improvement is not contingent on an increase in the number of model parameters but rather on CLIP's prowess in learning semantically rich visual representations. Furthermore, the inference time of the two remains equivalent, indicating that the incorporation of CLIP does not compromise the practical efficiency of the model. (2) We initially explore the CLIP-VIReID learning paradigm in this paper to fully harness the formidable potential of CLIP for VIReID. It can be seen that the TP has undergone an expansion. This primarily results from the requisite training of the prompt learner in the first stage and the attention pooling in the third stage. Specifically, the former comprises a TP of 0.8M, while the latter, a pivotal component within CLIP facilitating language-vision connections, encompasses approximately 15M TP. Despite this expansion, we contend it is acceptable, aligning with the prevailing trend in contemporary studies that employ large language-vision models for downstream tasks. (3)Building upon CLIP-VIReID, we propose CSDN to further facilitate VIReID learning. It can be seen that the modal-specific prompt learners increase TP in the first stage by 0.8M, and the attention fusion component in the second stage requires 4.2M TP. The performance gain from this small parameter increase is worth it. In addition, during the testing phase, we only employ the visual encoder to extract features for identity matching, and components from the first and second stages are unnecessary, leading to satisfactory inference times. In summary, the proposed method excels in performance without compromising practicality.

\subsubsection{Future Research}
In this paper, we pioneer the adaptation of CLIP to the VIReID task. However, this is only a preliminary exploration, and the utilization of superior language-vision models to align features between visible and infrared images warrants deeper investigation. In particular, we observe that language descriptions acquired through the prompt learner paradigm exhibit relative coarseness, primarily owing to the limited learnable word tokens, thus limiting the richness of semantic information. Consequently, our future research will focus on enriching semantic information linked to images, aiming to more effectively steer the learning of visual representations.

\section{Conclusion}
In this paper, we explore the application of CLIP in VIReID and propose a CLIP-Driven Semantic Discovery Network (CSDN) to facilitate learning of VIReID. Specifically, CSDN introduces bimodal learnable natural language descriptions to acquire the semantic information corresponding to visual representations of visible and infrared images. To promote the model to further sense rich complementary semantics, CSDN incorporates an attention fusion mechanism to integrate text features across different modalities. With the guidance of learnable natural language descriptions, CSDN injects semantic information into visual representations to improve their modality invariance. The proposed CSDN proficiently mitigates challenges arising from the modality gap in cross-modality matching. A series of comprehensive experiments and analyses unequivocally substantiate the effectiveness and superiority of the CSDN framework. Moving forward, we aim to further explore the potential of multi-modal large models on the VIReID task.


%





\ifCLASSOPTIONcaptionsoff
  \newpage
\fi



%
\bibliography{mybibfile}

\end{document}